\documentclass{ieeeaccess}
\usepackage{soul}

\usepackage{amssymb}
\usepackage{amsmath}
\usepackage{cite}
\usepackage{amsmath,amssymb,amsfonts}
\usepackage{algorithmic}
\usepackage{graphicx}
\usepackage{float}

\usepackage{tabularx,booktabs}
        \newcolumntype{L}{>{\raggedright\arraybackslash}X}
        \newcolumntype{C}{>{\centering\arraybackslash}X}

\usepackage[inline]{enumitem}
        
\usepackage{textcomp}
\def\BibTeX{{\rm B\kern-.05em{\sc i\kern-.025em b}\kern-.08em
    T\kern-.1667em\lower.7ex\hbox{E}\kern-.125emX}}

\usepackage{subfigure}
\usepackage{colortbl}
\usepackage{tipa}

\begin{document}

\history{Date of submission March 5, 2020}
\doi{10.1109/ACCESS.2017.DOI}

\title{Segmentation of Satellite Imagery using U-Net Models for Land Cover Classification}

\author{
\uppercase{Priit Ulmas}\authorrefmark{1},
\uppercase{Innar Liiv}\authorrefmark{1,2}
}


\address[1]{Department of Software Science, Tallinn University of Technology, Akadeemia Tee 15a, 12618 Tallinn, Estonia (e-mail: innar.liiv@taltech.ee, priit.ulmas@gmail.com)}
\address[2]{Centre for Technology and Global Affairs, University of Oxford, Manor Road, Oxford OX1 3UQ, United Kingdom}

\markboth
{Ulmas \headeretal: Segmentation of Satellite Imagery using U-Net Models for Land Cover Classification}
{Ulmas \headeretal: Segmentation of Satellite Imagery using U-Net Models for Land Cover Classification}

\corresp{Corresponding author: Priit Ulmas (e-mail: priit.ulmas@gmail.com).}

\begin{abstract}
The focus of this paper is using a convolutional machine learning model with a modified U-Net structure for creating land cover classification mapping based on satellite imagery. The aim of the research is to train and test convolutional models for automatic land cover mapping and to assess their usability in increasing land cover mapping accuracy and change detection. To solve these tasks, authors prepared a dataset and trained machine learning models for land cover classification and semantic segmentation from satellite images. The results were analysed on three different land classification levels. BigEarthNet satellite image archive was selected for the research as one of two main datasets. This novel and recent dataset was published in 2019 and includes Sentinel-2 satellite photos from 10 European countries made in 2017 and 2018. As a second dataset the authors composed an original set containing a Sentinel-2 image and a CORINE land cover map of Estonia. The developed classification model shows a high overall F\textsubscript{1} score of 0.749 on multiclass land cover classification with 43 possible image labels. The model also highlights noisy data in the BigEarthNet dataset, where images seem to have incorrect labels. The segmentation models offer a solution for generating automatic land cover mappings based on Sentinel-2 satellite images and show a high IoU score for land cover classes such as forests, inland waters and arable land. The models show a capability of increasing the accuracy of existing land classification maps and in land cover change detection.
\end{abstract}

\begin{keywords}
Satellite Imagery,
U-Net Models,
Land Cover Classification.
\end{keywords}

\titlepgskip=-15pt

\maketitle

\section{Introduction}
\label{sec:introduction}
\PARstart{W}{e} are witnessing rapid development in space technologies both in physical satellite deployments and in data processing capabilities. The growing number of earth observation (EO) satellites produce an expanding amount of data, which requires a set of tools supported by artificial intelligence to process and extract information from \cite{PHI2018}. This paper looks at deep convolutional neural networks (CNN) in a study of pixel level land cover classification mapping from satellite images.

Land cover mapping is a highly important tool for monitoring both the environmental development for regional planning as well as detecting changes in the environment. One such use case is monitoring UN Sustainable Development Goals (SDGs) \cite{SDG2015} \cite{ANDERSON2017}, specifically goal 15: Life on Land.

However, the current large-scale land cover maps have several weaknesses, most notably the complicated, labour intensive and time-consuming process of creating them. Preparing such maps often requires people from the local areas to validate and classify the data. And even though a lot of effort is put into creating the maps, they provide a relatively low spatial accuracy which is not sufficient for automated change detection of small-scale changes. The CORINE Land Cover (CLC) map, for example, has been created with a 6-year interval. The most recent one, 2018 version, had a production time of 1.5 years \cite{CORINE}.

When looking at ways of automatic change detection, for example in forestry mapping deforestation and illegal logging activity, a more precise and faster land cover mapping process is needed to detect and highlight small scale changes happening in weekly or even daily time frames.

Deep learning has shown high accuracy in computer vision tasks and has high potential to handle the growing amount of Earth Observation (EO) data in an automated process \cite{STORIE2018}. For generating land cover classification maps an image segmentation task needs to be solved.
Pixel level segmentation on satellite images is challenging because collecting a ground truth dataset for training such a segmentation model is difficult and time consuming.

Fortunately, a transfer learning approach can be taken, where a model trained for one task is repurposed to solve a new task. Specifically, in this work we use a large scale classification dataset (BigEarthNet) to develop a classification model and then repurpose this learning in a segmentation model.

The solution to this research problem is split into two tasks. Firstly, to create a land cover classification model using a large-scaled BigEarthNet dataset and secondly, to use this model as a pretrained encoder in a modified U-Net model to generate pixel level land cover classification maps, using a much smaller dataset for training. In the first task we benefit from the large dataset to learn the features of satellite images - the model will become good in reading satellite images. As in the second task we have a lot smaller dataset with more noise, we benefit from using a transfer learning approach to carry on the ability to read satellite images we trained in the first model.

The remainder of this paper is organized as follows. In Section \ref{sec:literature}, we introduce additional background and discuss the topics addressed in this paper. In Section \ref{sec:methodology}, we look at the approach taken in this work. In Section \ref{sec:experiments}, we see a summary of the machine learning training process and Section \ref{sec:results} describes the results. Finally, Section \ref{sec:conclusion} concludes the paper by summarizing the results and indicating issues to be addressed in future work.

\section{Related work}
\label{sec:literature}
The use of deep convolutional models has proven to deliver superior accuracy in a large variety of computer vision tasks and so also in satellite image understanding. One of the main bottlenecks in this approach has been the lack of labelled training data, which needs to be manually collected and prepared. In the case of image segmentation, the pixel level segmentation masks are even more difficult to collect. \cite{STORIE2018}

One way to overcome the lack of training data is to use a weakly supervised learning method. This has been recently employed by \cite{NIVAGGIOLI2019} and \cite{WANG2020}. Weakly supervised approaches aim to overcome the need for complex training datasets, which in many cases do not exist and are difficult to create. Nivaggioli et al. \cite{NIVAGGIOLI2019} used an approach suggested by \cite{AHN2018} on satellite imagery. Wang et al. \cite{WANG2020} explore weak labels in the form of a single pixel label per image and class activation maps to create pixel level land cover mappings.

Automated land cover mapping is an active field of research with many machine learning approaches suggested, solving tasks such as vegetation extraction \cite{ZHIYONG2019} or land cover change detection \cite{ZHAN2020}. In creating a global land cover map, CORINE \cite{BUCHHORN2019} uses several data fusion and pre-processing steps together with ancillary data sources to generate a training dataset. A rule-based approach combined with decision tree models is then applied to create a land cover map. To overcome the lack of large-scale image datasets for model training transfer learning and data augmentation methods are used \cite{SCOTT2017} \cite{BENBAHRIA2019}.

Different satellite data sources have been applied for automatic land cover detection. For land cover and crop type classification \cite{KUSSUL2017} used Sentinel-1A and Landsat-8 data. In \cite{GBODJO2019}, Sentinel-1 and Sentinel-2 data is used in a multi-source approach to benefit from combining radar and optical data as a time series.

In this work, firstly, a land cover classification model is created using the large-scale BigEarthNet dataset, followed by creating a modified U-Net model using a transfer learning approach. CORINE Land Cover data of 2018 is then used as a training set for the segmentation model and data augmentation is used in the model training process.

\section{Methodology}
\label{sec:methodology}
This chapter describes the main data pre-processing steps and the neural network architectures used.

\subsection{Data preparation}

Two data sources were used to train machine learning models. In the first stage BigEarthNet \cite{SUMBUL2019} was used to solve the classification task. Secondly, an original custom dataset was created to train the segmentation model for the semantic segmentation task.

\subsubsection{BigEarthNet dataset}

BigEarthNet is a large-scale Sentinel-2 dataset collected from a total of 125 Sentinel-2 tiles covering areas of 10 countries in Europe. The dataset was prepared with data from 2017. and 2018. and it was published in 2019. A total of 590 326 tiles of size 120 x 120px are annotated with land cover classification labels according to the third level of CORINE land cover classification covering a total of 43 different land cover classes. \cite{SUMBUL2019}

\Figure[!htbp](topskip=0pt, botskip=0pt, midskip=0pt)[width=0.99\columnwidth]{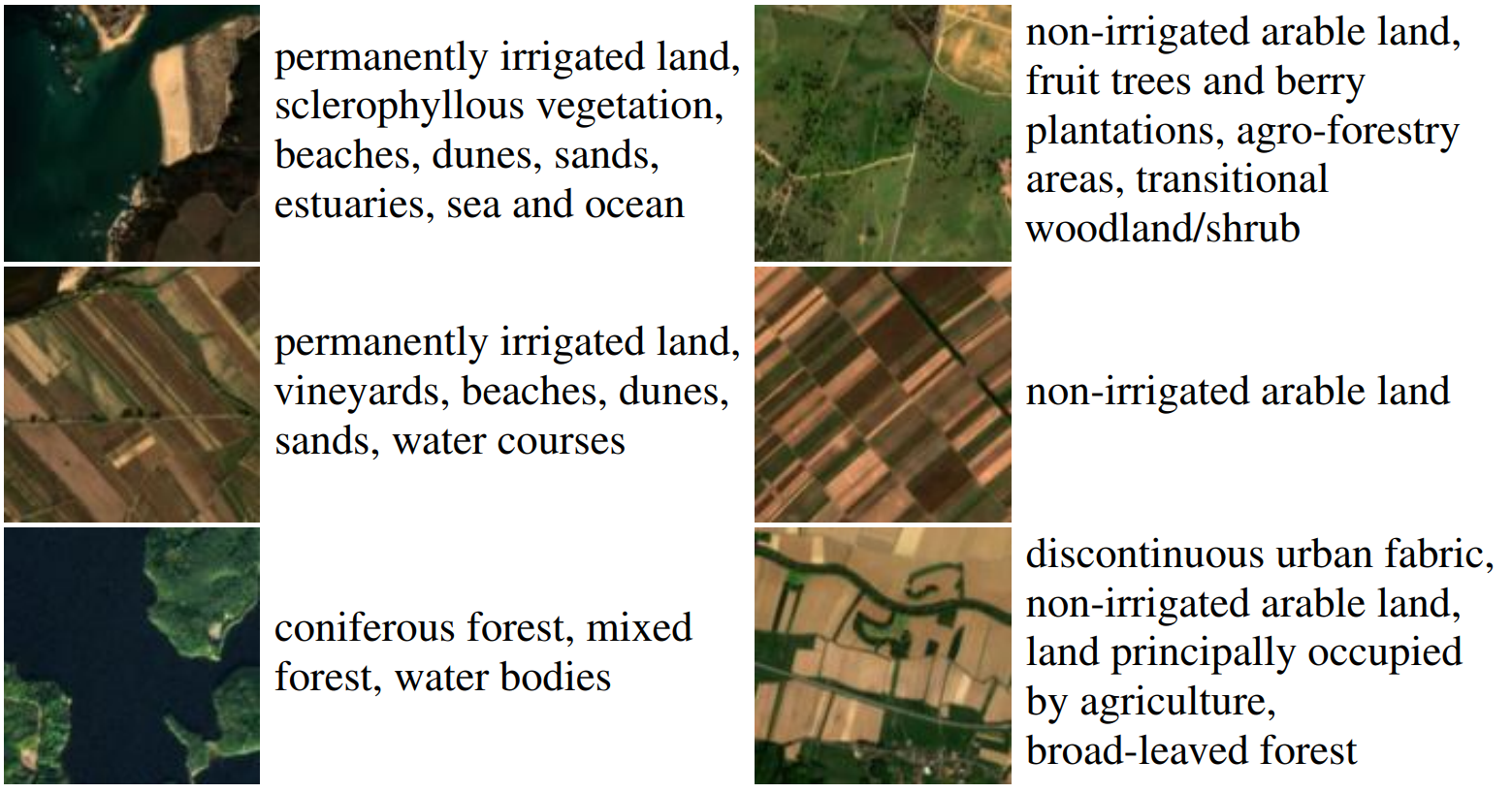}
{An example from the BigEarthNet dataset showing individual satellite images and corresponding image labels. \cite{SUMBUL2019} \label{bigearthnet_example}}

\begin{table}
    \renewcommand{\arraystretch}{1.03}
	\caption{CORINE Land cover classification labels. A three-level hierarchical classification structure is indicated by the first column.}
	\label{tab:CORINE_classification}
	\centering
    \begin{tabularx}{\linewidth}{|p{0.6cm}|L|p{0.9cm}|}
    \hline
    \textbf{Code} & \textbf{Land Cover Category} & \textbf{Images}  \\
    \hline
    1&Artificial surfaces&\\ \hline
    11&Urban fabric&\\ \hline
    111&Continuous urban fabric& 10 784\\ \hline
    112&Discontinuous urban fabric& 69 872\\ \hline
    12&Industrial, commercial and transport units&\\ \hline
    121&Industrial or commercial units& 12 895\\ \hline
    122&Road and rail networks and associated land& 3 384\\ \hline
    123&Port areas& 509\\ \hline
    124&Airports& 979\\ \hline
    13&Mine, dump and construction sites&\\ \hline
    131&Mineral extraction sites& 4 618\\ \hline
    132&Dump sites& 959\\ \hline
    133&Construction sites& 1 174\\ \hline
    14&Artificial, non-agricultural vegetated areas&\\ \hline
    141&Green urban areas& 1 786\\ \hline
    142&Sport and leisure facilities& 5 353\\ \hline
    2&Agricultural areas&\\ \hline
    21&Arable land&\\ \hline
    211&Non-irrigated arable land& 196 695\\ \hline
    212&Permanently irrigated land& 13 589\\ \hline
    213&Rice fields& 3 793\\ \hline
    22&Permanent crops&\\ \hline
    221&Vineyards& 9 567\\ \hline
    222&Fruit trees and berry plantations& 4 754\\ \hline
    223&Olive groves& 12 538\\ \hline
    23&Pastures&\\ \hline
    231&Pastures& 103 554\\ \hline
    24&Heterogeneous agricultural areas&\\ \hline
    241&Annual crops associated with permanent crops& 7 022\\ \hline
    242&Complex cultivation patterns& 107 786\\ \hline
    243&Land principally occupied by agriculture, with significant areas of natural vegetation& 147 095\\ \hline
    244&Agro-forestry areas& 30 674\\ \hline
    3&Forest and semi natural areas&\\ \hline
    31&Forests&\\ \hline
    311&Broad-leaved forest& 150 944\\ \hline
    312&Coniferous forest& 211 703\\ \hline
    313&Mixed forest& 217 119\\ \hline
    32&Scrub and/or herbaceous vegetation associations&\\ \hline
    321&Natural grasslands& 12 835\\ \hline
    322&Moors and heathland& 5 890\\ \hline
    323&Sclerophyllous vegetation& 11 241\\ \hline
    324&Transitional woodland-shrub& 173 506\\ \hline
    33&Open spaces with little or no vegetation&\\ \hline
    331&Beaches, dunes, sands& 1 578\\ \hline
    332&Bare rock& 3 277\\ \hline
    333&Sparsely vegetated areas& 1 563\\ \hline
    334&Burnt areas& 328\\ \hline
    4&Wetlands&\\ \hline
    41&Inland wetlands&\\ \hline
    411&Inland marshes& 6 236\\ \hline
    412&Peat bogs& 23 207\\ \hline
    42&Maritime wetlands&\\ \hline
    421&Salt marshes& 1 562\\ \hline
    422&Salines& 424\\ \hline
    423&Intertidal flats& 1 003\\ \hline
    5&Water bodies&\\ \hline
    51&Inland waters&\\ \hline
    511&Water courses& 10 572\\ \hline
    512&Water bodies& 83 811\\ \hline
    52&Marine waters&\\ \hline
    521&Coastal lagoons& 1 498\\ \hline
    522&Estuaries& 1 086\\ \hline
    523&Sea and ocean& 81 612\\ \hline
    \end{tabularx}
\end{table}

In pre-processing a total of 70 987 images with cloud coverage or snow \cite{SUMBUL2019} were left out of the data and the remaining was split into training and test sets. An 80\% subset of the data was used for training the classification model and the remaining 20\% set was used for validation.

Additionally, the classification labels were formatted into three levels of classification, corresponding to the hierarchy of CORINE land cover classification (Table \ref{tab:CORINE_classification}). \cite{BUTTNER2004}

\subsubsection{Sentinel-2 and CORINE combined dataset}

To train the segmentation model a custom dataset was combined from the CORINE Land Cover map (2018) and a Sentinel-2 satellite image. For the purposes of this research a single Sentinel-2 image was selected together with the corresponding land cover mapping (Fig.~\ref{estonia}). Sentinel-2 tile: S2A\_MSIL1C\_20180510T094031\_\-N0206\_\-R036\_\-T35VMF\_\-20180510T114819

\Figure[!htbp](topskip=0pt, botskip=0pt, midskip=0pt)[width=0.99\columnwidth]{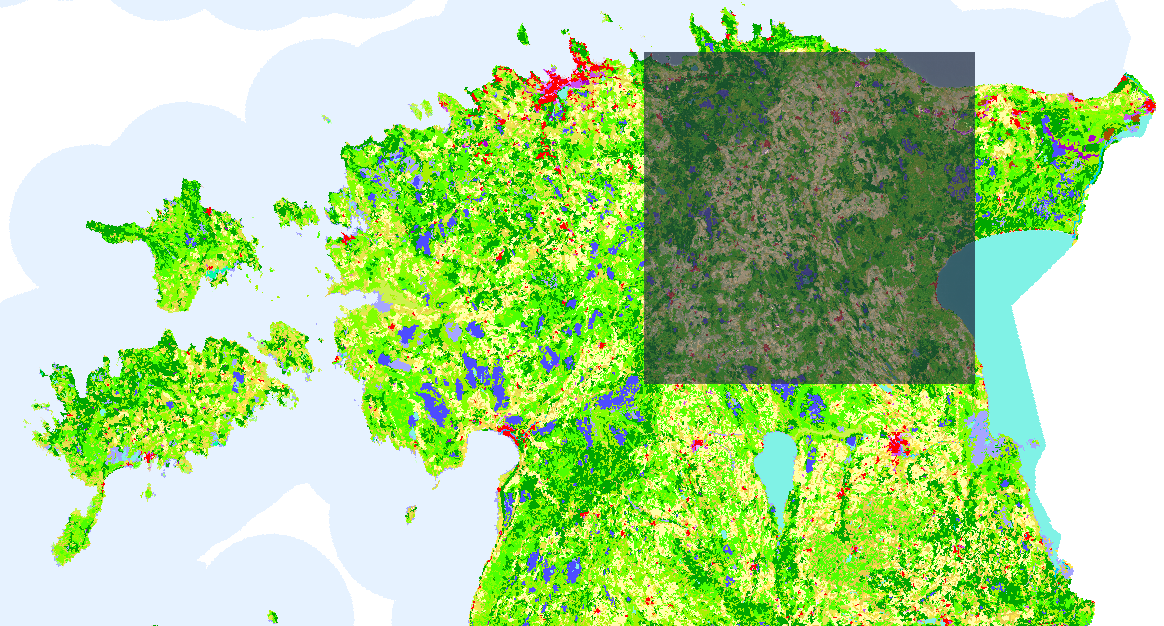}
{Square indicates the satellite image used for training the segmentation model. \label{estonia}}

The R, G, B channels of the satellite image were used in the case study, combined into a single .png image. Both the satellite image and the land cover map were then divided into 120 x 120px images, resulting in a dataset of 8281 image pairs.

The CORINE Land Cover classification has a three-level hierarchical structure \cite{BUTTNER2004} as shown in Table~\ref{tab:CORINE_classification}. The segmentation labels were also formatted into three separate sets corresponding to the hierarchy of land cover classification labels.

The created dataset is a small sample of satellite data and accounts for a smaller set of labels compared to the classification dataset. Where BigEarthNet has 5, 15 and 43 different land cover classes over three classification levels, the segmentation dataset has 5, 14 and 25 respectively. The main difference is therefore on the third level, where fewer classes are present.

An 80\% subset of the data was used for training the segmentation model and the remaining 20\% set was kept for validation. Understandably the much lower precision of the CORINE Land Cover map introduces noise into the segmentation dataset when raised to the higher resolution of a Sentinel-2 satellite image. This will also impact the accuracy results as a portion of this dataset is used for validation.

\subsubsection{Dataset accuracy}
The accuracy of the CORINE land cover map used in the segmentation dataset is estimated to be close to 85\% \cite{2017CORINEREPORT}. The data accuracy is also defined by its Minimum Mapping Unit (MMU) of 25 ha for areas and 100 m for linear instances \cite{BUTTNER2004}, resulting in a relatively high level of generalisation. This omitting of small features creates a noisy dataset for the convolutional model to train on and means that there is no high accuracy ground truth data to rely on or to measure model results on.

\subsection{Network architecture}
Two neural network architectures were used in the case study. Firstly, a ResNet50 model \cite{HE2016} was used for the classification task. Secondly, the pretrained ResNet50 classification model was used as the encoder in the modified U-Net model \cite{RONNEBERGER2015} to solve the segmentation task. This transfer learning approach \cite{TAN2018} allows us to use the learning gained in the first task to solve a new, more complex task where training data is difficult to acquire. In the case of traditional image recognition tasks, this might mean that a model previously trained on ImageNet \cite{RUSSAKOVSKY2015} is used as a starting point. This model, already trained on a dataset of over 14 million images, is good at identifying different features on an image. To solve a new task only the final layers might need to be changed and trained in order to generate the required output at high accuracy.

\subsubsection{Classification model}
For the classification task a ResNet model architecture was chosen. This architecture uses a repeating pattern of layer blocks, with skip connections added to allow creating deeper networks while avoiding model performance degrading. This state-of-the-art model architecture also won the ImageNet Large Scale Visual Recognition Challenge (ILSVRC) in image classification in 2015. \cite{HE2016}

\subsubsection{Segmentation model}
For image segmentation a U-Net like model architecture is used. This type of model is comprised of two main parts. Firstly, the encoder half of the model is used to detect features on an image. This portion of the model carries out a downsampling process, bringing the input image down to a small size feature matrix. Secondly, the decoder half constructs the model output using the features as input and carries out an upsampling process to bring back the spatial information of the input image.

\subsection{Evaluation metrics}
In both tasks the model will be assessed based on a validation set made up of 20\% of the data (103 867 images in classification and 1 656 images in segmentation). The following metrics are used in assessing the model results, described by \cite{SOROWER2010} and \cite{BERNARDINI2013}.

To analyse the BigEarthNet dataset to understand its complexity two metrics are used. Firstly, cardinality shows the average number of labels per image.

\begin{equation} \label{cardinality}
    Cardinality=\frac{1}{N}\sum_{i=1}^{N}|Y_{i}|
\end{equation}

Secondly, density is a metric that shows the average number of image labels out of all possible labels in the dataset.

\begin{equation} \label{density}
    Density=\frac{1}{N}\sum_{i=1}^{N}\frac{|Y_{i}|}{|L|}\qquad where \quad |L|=\bigcup_{i=1}^N Y_{i}
\end{equation}

The strictest measure used for classification is Exact Match Ratio (MR) (\ref{match_ratio}), where only the images with all labels correctly predicted are considered correct. It is strict because partially correct results will be considered incorrect.

\begin{equation} 
    \label{match_ratio}
    MR=\frac{1}{N}\sum_{i=1}^{N}I({Y_{i}=Z_{i}})
\end{equation}

Precision metric (\ref{precision}) is used to see the rate of correct labels out of all predicted labels. It combines the True Positive (TP) and False Positive (FP) results:

\begin{equation}
    \label{precision}
    Precision=\frac{TP}{TP+FP}
\end{equation}

The recall metric (\ref{recall}) is used to measure the proportion of correct labels out of all predicted labels. It combines the True Positive (TP) and False Negative (FN) results:

\begin{equation}
    \label{recall}
    Recall=\frac{TP}{TP+FN}
\end{equation}

For the classification task the F\textsubscript{1} value (\ref{fbeta}) will be used for model assessment. The F-score combines precision and recall into a single metric. In the case of F\textsubscript{1} we are putting an equal weight on both of these metrics.

\begin{equation}
    \label{fbeta}
    F_1=\frac{2*precision*recall}{(precision)+recall}
\end{equation}

Segmentation accuracy is measured using overall accuracy (\ref{seg_tapsus}) as well by using class based Jaccard Index values (\ref{jaccard}).

\begin{equation} 
    \label{seg_tapsus}
    Segmentation\ accuracy=\frac{Correct\ pixels}{All\ pixels}
\end{equation}

\begin{equation} 
    \label{jaccard}
    Jaccard\ Index=\frac{TP}{TP+FN+FP}
\end{equation}

\section{Experiments}
\label{sec:experiments}
\subsection{Training environment}
The machine learning models used in the research were prepared using the Fast.ai library \cite{HOWARD2020}. Built on the Pytorch framework \cite{PASZKE2019}, this high-level library is created with the aim of simplifying state-of-the-art model creation in deep learning. For the current work it enables creating a U-Net like architecture from an existing convolutional model, such as ResNet. The training was carried out on a virtual machine using Nvidia K80 and P100 graphics cards.

\subsection{Model training}
In the experiments a total of six models were trained and their results analysed. Firstly, three classification models were created, one for each level of CORINE Land Cover classification. The BigEarthNet dataset was used in training the models. Secondly, the three classification models were used as pretrained encoders for three U-Net like segmentation models. These models were then trained on the dataset created by the authors.

\subsubsection{Classification model training}
An ImageNet \cite{RUSSAKOVSKY2015} pretrained ResNet50 model was used to create the classification models. A total of three classification models were trained, one for each land cover classification level. 

The model was trained on the BigEarthNet dataset, with satellite images as input and land classification labels as output. As the model was pretrained at start, the training was carried out in two stages, firstly by keeping most of the layers frozen and only training the last layers, later by training the whole model.

The training was carried out in a total of 15 epochs (10 epochs training only the last layers of the model followed by 5 epochs with all layers unfrozen).

\subsubsection{Segmentation model training}
For the segmentation task three modified U-Net models were created using the previously trained classification model weights as pre-trained encoders. These models were then trained on the custom land cover segmentation dataset, taking as input a satellite image and as output a CORINE Land Cover map.

As the models are using a pretrained model as the encoder the training is carried out in two stages. Firstly, only the decoder is trained (5-10 epochs), followed by unfreezing all layers and training the whole model (5-10 epochs). A similar process is carried out for all three classification levels.

\section{Results}
\label{sec:results}
\subsection{Dataset analysis}

In analysis of BigEarthNet through the three levels of CORINE Land Cover classification two metrics were calculated - cardinality and density. These characteristics allow us to illustrate the multi-class and multi-label dataset. Results are shown in table \ref{tab:cardinality&density}.

\begin{table}[!htbp]
    \renewcommand{\arraystretch}{1.3}
    \centering
	\caption{Cardinality and density of the BigEarthNet dataset.}
    \begin{tabular}{|c|c|c|}
        \hline
    	\textbf{Classification level} & \textbf{Cardinality} & \textbf{Density}   \\
    	\hline
        Level 1&1.78&0.357\\ \hline
        Level 2&2.48&0.165\\ \hline
        Level 3&2.96&0.069\\ \hline
    \end{tabular}
    \label{tab:cardinality&density}
\end{table}

From the results we can see that from the first to the third level of classification the cardinality rises from 1.78 to 2.96, meaning that between the first and third levels the average number of labels for an image roughly doubles. However, from density the complexity rises more, choosing the labels out of all possible choices becomes nearly five times more difficult (cardinality reducing from 0.357 on the first level down to 0.069 on the third level).

Both of the training datasets are strongly imbalanced on all classification levels, meaning that there are a few classes highly represented followed by many classes with less data. One question explored in the work was to understand if the higher representation of a class is resulting in a higher score for the class. When looking at the correlation between the number of pictures per class and the class based F\textsubscript{1} score we can see a correlation value of \textbf{0.59}, indicating a medium correlation between the two. On the segmentation task we looked at the total number of pixels per class and the IoU score of the class. In this case we also see a medium correlation value of \textbf{0.66}. This shows that in addition to the class being represented in the dataset the visual distinction of the class plays an important role.

\subsection{Overall results}

\subsubsection{Classification}
The classification model shows high overall results (Table~\ref{results}) on all three land cover classification levels. While the exact match ratio, where all labels are correctly predicted, drops from 75.3\% on the first level down to 33.1\% on the third level the proportion of images where all predicted labels are incorrect rises from 0.8\% to only 4.9\% on the third level. F\textsubscript{1} score remains relatively high on all three levels, going down to 0.749 for the third level. This F\textsubscript{1} score on the third level is also an improvement over the F\textsubscript{1} score of 0.6759 reached by \cite{SUMBUL2019}.

\begin{table}[!htbp]
    \renewcommand{\arraystretch}{1.3}
    \caption{Classification and segmentation results. Exact match ratio (MR) indicates complete accurate classification, Partial indicates images with partially correct predictions and Incorrect indicates images with no correct predicted labels.}
    \label{results}
    \centering
    \begin{tabular}{|c|c|c|c|c|c|}
        \hline
        {} & \multicolumn{4}{c|}{\textbf{Classification}} & \textbf{Segmentation}\\
        \hline
        & \textbf{MR} & \textbf{Partial} & \textbf{Incorrect} & \textbf{F\textsubscript{1}} & \textbf{Accuracy}\\
        \hline
        Level 1 & 75.3\% & 23.9\% & 0.8\% & 0.920 & 91.4\% \\
        \hline
        Level 2 & 45.9\% & 52.1\% & 2.1\% & 0.823 & 75.8\% \\
        \hline
        Level 3 & 33.1\% & 61.9\% & 4.9\% & 0.749 & 59.7\% \\
        \hline
    \end{tabular}
\end{table}

\subsubsection{Segmentation}

The segmentation model shows a high 91.4\% pixel-level accuracy on the first classification level with 75.8\%  and 59.7\% on the second and third levels (Table~\ref{results}). An example of segmentation model results and comparison to validation data can be seen on Figures \ref{seg_lvl1_comparison} and \ref{seg_example}.

\Figure[!htbp](topskip=0pt, botskip=0pt, midskip=0pt)[width=0.95\columnwidth]{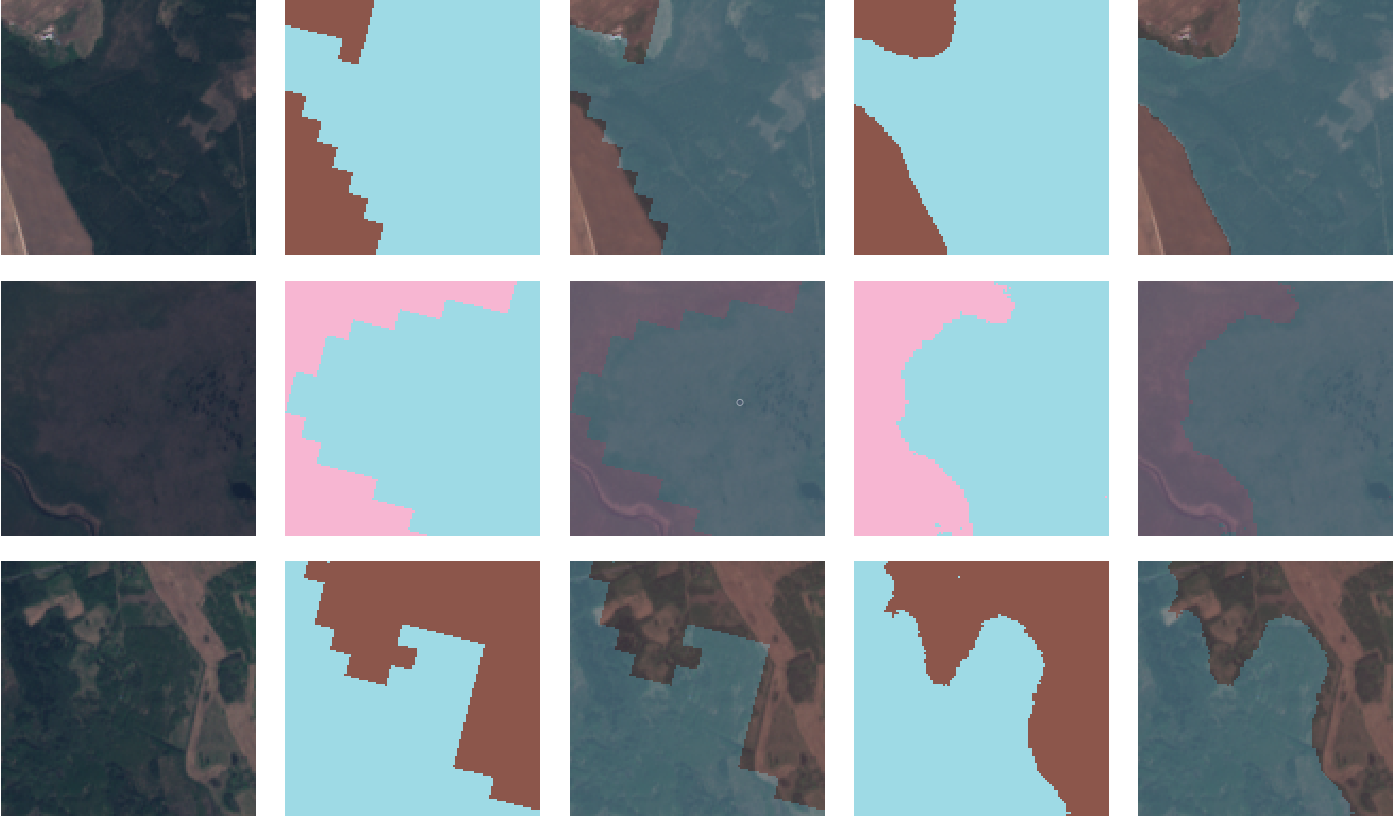}
{Segmentation results comparison. Satellite image tiles (column 1), existing land cover maps (column 2), segmentation results (column 4) and land cover map overlays (columns 3 and 5). \label{seg_lvl1_comparison}}

\Figure[!htbp](topskip=0pt, botskip=0pt, midskip=0pt)[width=0.99\columnwidth]{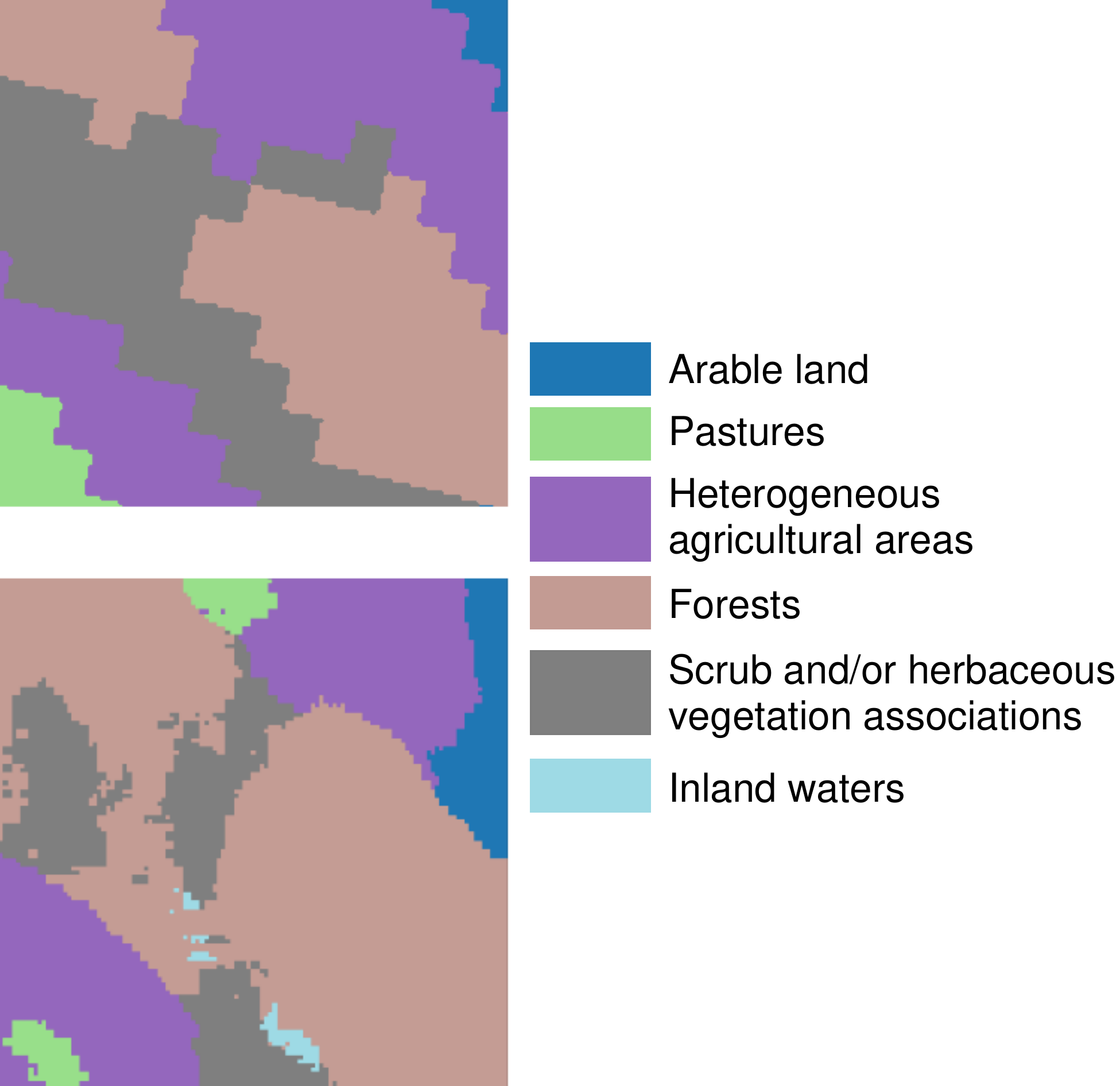}
{Example of segmentation results on the third level of classification (below) compared to the CORINE Land Cover map used in the validation set (above).
\label{seg_example}}

\subsection{Class based accuracy}

\subsubsection{Classification}
The classification results per class show high metrics on the first level and lower scores on the second and third levels, where the number of classes increases and visual separability of classes decreases.

On the first level (Table~\ref{lvl1_class_results}) we can see that the model preforms the weakest on wetland areas (F\textsubscript{1} score 0.58) and artificial surfaces (F\textsubscript{1} score 0.77).

\begin{table}[!htbp]
    \renewcommand{\arraystretch}{1.3}
    \caption{Classification level 1 results.}
    \label{lvl1_class_results}
    \centering
    \begin{tabularx}{\columnwidth}{|L|c|c|c|c|}
        \hline
        \textbf{Class} & \textbf{Precision} & \textbf{Recall}  & \textbf{F\textsubscript{1}} & \textbf{Images} \\
        \hline
        Agricultural areas & 0.95 & 0.93 & 0.94 & 64 182 \\
        \hline
        Artificial surfaces & 0.85 & 0.71 & 0.77 & 17 689 \\
        \hline
        Forests & 0.94 & 0.95 & 0.95 & 70 138 \\
        \hline
        Water bodies & 0.95 & 0.91 & 0.93 & 28 570 \\
        \hline
        Wetlands & 0.85 & 0.45 & 0.58 & 4 869 \\
        \hline
    \end{tabularx}
\end{table}

On the second level of classification (Table~\ref{lvl2_class_results}) we can see that most classes still show F\textsubscript{1} score above 0.5, with lower values only for industrial areas and scrub and/or herbaceous vegetation classes. These classes both have a small number of images and both can be difficult for the model to visually distinguish from other classes.

\begin{table}[!htbp]
    \renewcommand{\arraystretch}{1.3}
    \caption{Classification level 2 results.}
    \label{lvl2_class_results}
    \centering
    \begin{tabularx}{\columnwidth}{|L|c|c|c|c|}
        \hline
        \textbf{Class} & \textbf{Precision} & \textbf{Recall}  & \textbf{F\textsubscript{1}} & \textbf{Images} \\
        \hline
        Urban fabric&0.87&0.85&0.86&38 755 \\ \hline
        Industrial, commercial, transport&0.74&0.11&0.20&1 299 \\ \hline
        Mine, dump, construction&0.93&0.94&0.94&64 364 \\ \hline
        Artificial, non-agri. vegetated areas&0.83&0.77&0.80&42 671 \\ \hline
        Arable land&0.73&0.44&0.55&3 050 \\ \hline
        Permanent crops&0.90&0.79&0.84&13 542 \\ \hline
        Pastures&0.84&0.45&0.58&4 424 \\ \hline
        Heterogeneous agricult. areas&0.99&0.98&0.98&15 083 \\ \hline
        Forests&0.84&0.42&0.56&455 \\ \hline
        Scrub and/or herbaceous vegetation&0.81&0.16&0.26&1 197 \\ \hline
        Open spaces, little or no vegetation&0.74&0.44&0.55&1 199 \\ \hline
        Inland wetlands&0.83&0.67&0.74&19 867 \\ \hline
        Maritime wetlands&0.75&0.43&0.54&4 968 \\ \hline
        Inland waters&0.75&0.59&0.66&31 820 \\ \hline
        Marine waters&0.83&0.72&0.77&15 118 \\ \hline
    \end{tabularx}
\end{table}

Looking at the third level of classification (Table~\ref{lvl3_class_results}) we see that as the variety of classes increased nearly 3x between the second and third level the range of scores has spread to some classes missed completely (mineral extraction sites, sport and leisure facilities) and others reaching nearly maximum score (salt marshes, F\textsubscript{1} score 0.98).

\begin{table}[!htbp]
    \renewcommand{\arraystretch}{1.3}
    \caption{Classification level 3 results.}
    \label{lvl3_class_results}
    \centering
    \begin{tabularx}{\columnwidth}{|L|c|c|c|c|}
        \hline
        \textbf{Class} & \textbf{Precision} & \textbf{Recall}  & \textbf{F\textsubscript{1}} & \textbf{Images} \\ \hline
        Continuous urban fabric&0.81&0.76&0.78&6065 \\ \hline
        Discontinuous urban fabric&0.50&0.01&0.01&172 \\ \hline
        Industrial or commercial units&0.62&0.26&0.36&1408 \\ \hline
        Road and rail networks and associated land&0.73&0.52&0.61&549 \\ \hline
        Port areas&0.82&0.57&0.68&359 \\ \hline
        Airports&0.80&0.72&0.76&28 253 \\ \hline
        Mineral extraction sites&0.00&0.00&0.00&72 \\ \hline
        Dump sites&0.86&0.72&0.78&302 \\ \hline
        Construction sites&0.75&0.64&0.69&20 948 \\ \hline
        Green urban areas&0.86&0.86&0.86&32 765 \\ \hline
        Sport and leisure facilities&0.00&0.00&0.00&247 \\ \hline
        Non-irrigated arable land&0.80&0.67&0.73&2 255 \\ \hline
        Permanently irrigated land&0.83&0.69&0.76&13 238 \\ \hline
        Rice fields&0.83&0.07&0.12&153 \\ \hline
        Vineyards&0.70&0.45&0.55&223 \\ \hline
        Fruit trees and berry plantations&0.79&0.16&0.27&922 \\ \hline
        Olive groves&0.61&0.07&0.13&319 \\ \hline
        Pastures&0.72&0.39&0.51&2 410 \\ \hline
        Annual crops associated with permanent crops&0.83&0.10&0.17&1 079 \\ \hline
        Complex cultivation patterns&0.82&0.33&0.47&193 \\ \hline
        Land principally occupied by agriculture, with significant areas of natural vegetation&0.79&0.50&0.61&26 257 \\ \hline
        Agro-forestry areas&0.78&0.14&0.23&819 \\ \hline
        Broad-leaved forest&0.81&0.82&0.81&35 380 \\ \hline
        Coniferous forest&0.77&0.20&0.32&1 050 \\ \hline
        Mixed forest&0.76&0.31&0.44&2 226 \\ \hline
        Natural grasslands&0.87&0.82&0.84&36 743 \\ \hline
        Moors and heathland&0.73&0.29&0.42&2 489 \\ \hline
        Sclerophyllous vegetation&0.84&0.66&0.74&19 867 \\ \hline
        Transitional woodland-shrub&0.84&0.54&0.65&3 363 \\ \hline
        Beaches, dunes, sands&0.77&0.50&0.61&2 683 \\ \hline
        Bare rocks&0.68&0.30&0.41&88 \\ \hline
        Sparsely vegetated areas&0.68&0.49&0.57&725 \\ \hline
        Burnt areas&0.60&0.06&0.11&652 \\ \hline
        Glaciers and perpetual snow&0.76&0.43&0.55&90 \\ \hline
        Inland marshes&0.80&0.33&0.46&286 \\ \hline
        Peat bogs&0.82&0.38&0.52&2 218 \\ \hline
        Salt marshes&0.99&0.98&0.98&14 603 \\ \hline
        Salines&0.66&0.10&0.17&248 \\ \hline
        Intertidal flats&0.77&0.07&0.13&1 007 \\ \hline
        Water courses&0.73&0.55&0.63&29 479 \\ \hline
        Water bodies&0.77&0.25&0.38&1 889 \\ \hline
        Coastal lagoons&0.90&0.76&0.83&11 706 \\ \hline
        Estuaries&0.81&0.65&0.72&1 938 \\ \hline
        Sea and ocean&0.81&0.65&0.72&1 938 \\ \hline
    \end{tabularx}
\end{table}

\subsubsection{Segmentation}

Segmentation results at the first level of land cover classification show a high accuracy for the main land cover classes. This is also supported by visual comparison to the land cover map data used for validation (Figure~\ref{seg_lvl1_comparison}).

On the first level of classification we can see from the confusion matrix (Figure~\ref{seg_lvl1_confusion}) that the segmentation model performs well on three out of five classes.

\begin{table}[!htbp]
    \renewcommand{\arraystretch}{1.3}
    \caption{Segmentation Confusion Matrix for the first classification level. Rows indicate true labels and columns indicate model predictions. Main diagonal shows class based correct prediction accuracy.}
    \centering
    \label{seg_lvl1_confusion}
    \begin{tabularx}{0.8\columnwidth}{@{}l*{4}{C}c@{}}
        \toprule
        Labels &1&2&3&4&5\\
        \midrule
        1. Artificial surfaces&\cellcolor[gray]{.83}\textbf{0.42}&\cellcolor[gray]{.83}0.42&\cellcolor[gray]{.94}0.16&&\\
        2. Agricultural areas&\cellcolor[gray]{.99}0.02&\cellcolor[gray]{.69}\textbf{0.77}&\cellcolor[gray]{.92}0.21&&\\
        3. Forests&0.01&\cellcolor[gray]{.96}0.11&\cellcolor[gray]{.66}\textbf{0.86}&\cellcolor[gray]{.99}0.02&\\
        4. Wetlands&&0.01&\cellcolor[gray]{.77}0.57&\cellcolor[gray]{.84}\textbf{0.40}&0.01\\
        5. Water bodies&&\cellcolor[gray]{.98}0.05&\cellcolor[gray]{.96}0.09&\cellcolor[gray]{.98}0.04&\cellcolor[gray]{.67}\textbf{0.83}\\
        \bottomrule
    \end{tabularx}
\end{table}

On the second classification level (Figure~\ref{seg_lvl2_confusion}) we can see that the strongest results come from arable land, forest and inland waters classes. From one side these are among the most represented classes by pixel count, but they are also visually distinct. Looking at the other classes we can see that many of them are visually not separable from RGB images and the results spread among visually similar classes.

\begin{table*}[!htbp]
    \renewcommand{\arraystretch}{1.3}
    \caption{Segmentation Confusion Matrix for the second classification level. Rows indicate true labels and columns indicate model predictions. Main diagonal shows class based correct prediction accuracy.}
    \centering
    \label{seg_lvl2_confusion}
    \begin{tabularx}{\textwidth}{@{}l*{15}{C}}
        \toprule
        Labels&11&12&13&14&21&22&23&24&31&32&33&41&42&51&52 \\
        \midrule
        11. Urban fabric&\cellcolor[gray]{.83}\textbf{0.42}&\cellcolor[gray]{.99}0.03&&&\cellcolor[gray]{.93}0.17&&&\cellcolor[gray]{.88}0.29&\cellcolor[gray]{.96}0.09&&&&&&\\
        12. Industrial, commercial, transport&\cellcolor[gray]{.85}0.38&\cellcolor[gray]{.93}\textbf{0.18}&\cellcolor[gray]{.99}0.02&&\cellcolor[gray]{.92}0.19&&&\cellcolor[gray]{.94}0.14&\cellcolor[gray]{.97}0.08&0.01&&&&&\\
        13. Mine, dump, construction&&\cellcolor[gray]{.98}0.06&\cellcolor[gray]{.95}\textbf{0.13}&&\cellcolor[gray]{.94}0.16&&\cellcolor[gray]{.99}0.03&\cellcolor[gray]{.97}0.08&\cellcolor[gray]{.88}0.31&\cellcolor[gray]{.92}0.2&&0.01&&&\cellcolor[gray]{.99}0.03\\
        14. Artificial, non-agricultural vegetated areas&\cellcolor[gray]{.86}0.34&\cellcolor[gray]{.98}0.06&\cellcolor[gray]{.97}0.07&\textbf{0}&&&&\cellcolor[gray]{.96}0.11&\cellcolor[gray]{.84}0.4&0.01&&&&&\\
        21. Arable land&&&&&\textbf{0.76}&&\cellcolor[gray]{.97}0.07&\cellcolor[gray]{.97}0.05&\cellcolor[gray]{.95}0.12&&&&&&\\
        \addlinespace
        22. Permanent crops&&&&&\cellcolor[gray]{.97}0.07&\textbf{0}&\cellcolor[gray]{.87}0.32&\cellcolor[gray]{.94}0.16&\cellcolor[gray]{.88}0.29&\cellcolor[gray]{.94}0.15&&&&&\\
        23. Pastures&0.01&&&&\cellcolor[gray]{.86}0.36&&\cellcolor[gray]{.89}\textbf{0.28}&\cellcolor[gray]{.96}0.11&\cellcolor[gray]{.91}0.22&0.01&&&&&\\
        24. Heterogeneous agricult. areas&\cellcolor[gray]{.99}0.02&&&&\cellcolor[gray]{.88}0.29&&\cellcolor[gray]{.97}0.07&\cellcolor[gray]{.88}\textbf{0.3}&\cellcolor[gray]{.88}0.3&0.01&&&&&\\
        31. Forests&&&&&\cellcolor[gray]{.98}0.05&&0.01&\cellcolor[gray]{.99}0.03&\cellcolor[gray]{.65}\textbf{0.87}&\cellcolor[gray]{.99}0.03&&0.01&&&\\
        32. Scrub and/or herbaceous vegetation&0.01&&&&\cellcolor[gray]{.98}0.04&&\cellcolor[gray]{.99}0.02&\cellcolor[gray]{.98}0.05&\cellcolor[gray]{.73}0.67&\cellcolor[gray]{.94}\textbf{0.16}&&\cellcolor[gray]{.98}0.05&&&\\
        \addlinespace
        33. Open spaces, little or no vegetation&&&&&&&&&\cellcolor[gray]{.87}0.32&&\textbf{0}&&&\cellcolor[gray]{.73}0.68&\\
        41. Inland wetlands&0.01&&&&0.01&&&0.01&\cellcolor[gray]{.87}0.32&\cellcolor[gray]{.93}0.18&&\cellcolor[gray]{.82}\textbf{0.46}&&0.01&\\
        42. Maritime wetlands&&&&&&&&&&&&&\textbf{0}&&\\
        51. Inland waters&&&&&&&&\cellcolor[gray]{.98}0.04&\cellcolor[gray]{.96}0.1&&&\cellcolor[gray]{.98}0.05&&\cellcolor[gray]{.68}\textbf{0.8}&\\
        52. Marine waters&&&&&0.01&&0.01&\cellcolor[gray]{.98}0.04&\cellcolor[gray]{.98}0.05&&&\cellcolor[gray]{.99}0.02&&\cellcolor[gray]{.8}0.51&\cellcolor[gray]{.86}\textbf{0.36}\\
        \bottomrule
    \end{tabularx}
\end{table*}

The third level has the largest number of classes (25 are present in the segmentation dataset used for training and validation) and shows a wide distribution of results (Figure~\ref{seg_lvl3_confusion}). The best performing classes are also understandably similar to the first two levels. Arable land, forests, water bodies and urban fabric being the highest performing classes. From the confusion matrix we can also see that some of the error is happening between bordering areas, such as urban fabric and green urban areas. This error to some extent was introduced into the dataset by using the CORINE Land Cover map at the higher resolution of satellite images. This type of error is described next.

\begin{table*}[!htbp]
    \scriptsize
    \renewcommand{\arraystretch}{1.3}
    \caption{Segmentation Confusion Matrix for the third classification level. Rows indicate true labels and columns indicate model predictions. Main diagonal shows class based correct prediction accuracy. Class labels as Table \ref{tab:CORINE_classification}.}
    \centering
    \label{seg_lvl3_confusion}
    \begin{tabularx}{\textwidth}{@{}l*{25}{C}}
        \toprule
        Labels&111&121&122&123&124&131&132&141&142&211&222&231&242&243&311&312&313&321&322&324&331&411&412&512&523 \\
        \midrule
        111&\cellcolor[gray]{.81}\textbf{0.48}&\cellcolor[gray]{.99}0.02&&&&&&&&\cellcolor[gray]{.92}0.2&&&0.01&\cellcolor[gray]{.92}0.21&&\cellcolor[gray]{.99}0.02&\cellcolor[gray]{.98}0.05&&&0.01&&&&&\\
        121&\cellcolor[gray]{.79}0.52&\cellcolor[gray]{.96}\textbf{0.11}&&&&\cellcolor[gray]{.99}0.02&&&&\cellcolor[gray]{.91}0.22&&&&\cellcolor[gray]{.97}0.08&&0.01&\cellcolor[gray]{.99}0.03&&&0.01&&&&&\\
        122&\cellcolor[gray]{.94}0.14&0.01&\cellcolor[gray]{.98}\textbf{0.06}&&&0.01&&&&\cellcolor[gray]{.92}0.21&&\cellcolor[gray]{.99}0.03&&\cellcolor[gray]{.91}0.23&\cellcolor[gray]{.96}0.11&\cellcolor[gray]{.97}0.07&\cellcolor[gray]{.95}0.12&&&0.01&&&&&\\
        123&&&&\textbf{0}&&&&&&&&&&&&&&&&&&&&&\\
        124&\cellcolor[gray]{.90}0.25&\cellcolor[gray]{.84}0.4&&&\textbf{0}&\cellcolor[gray]{.98}0.06&&&&\cellcolor[gray]{.97}0.07&&&&\cellcolor[gray]{.94}0.16&&&0.01&&&\cellcolor[gray]{.98}0.04&&&&&\\
        \addlinespace
        131&\cellcolor[gray]{.99}0.02&\cellcolor[gray]{.98}0.04&&&&\cellcolor[gray]{.97}\textbf{0.08}&&&&\cellcolor[gray]{.88}0.29&&\cellcolor[gray]{.99}0.02&&\cellcolor[gray]{.98}0.05&&\cellcolor[gray]{.96}0.1&\cellcolor[gray]{.96}0.09&&&\cellcolor[gray]{.90}0.26&&&&&\cellcolor[gray]{.98}0.06\\
        132&\cellcolor[gray]{.93}0.18&\cellcolor[gray]{.96}0.1&&&&\cellcolor[gray]{.96}0.11&\textbf{0}&&&\cellcolor[gray]{.94}0.16&&&&\cellcolor[gray]{.98}0.05&&\cellcolor[gray]{.95}0.13&\cellcolor[gray]{.98}0.04&&&\cellcolor[gray]{.95}0.13&&\cellcolor[gray]{.98}0.06&&\cellcolor[gray]{.99}0.03&\\
        141&\cellcolor[gray]{.8}0.5&&&&&&&\textbf{0}&&&&&&&&\cellcolor[gray]{.8}0.5&&&&&&&&&\\
        142&\cellcolor[gray]{.94}0.14&\cellcolor[gray]{.88}0.31&&&&0.01&&&\textbf{0}&&&&&\cellcolor[gray]{.8}0.51&&&&&&\cellcolor[gray]{.99}0.02&&&&&\\
        211&0.01&&&&&&&&&\cellcolor[gray]{.68}\textbf{0.79}&&\cellcolor[gray]{.97}0.08&&\cellcolor[gray]{.99}0.03&0.01&0.01&\cellcolor[gray]{.97}0.07&&&&&&&&\\
        \addlinespace
        222&&&&&&&&&&\cellcolor[gray]{.94}0.15&\textbf{0}&\cellcolor[gray]{.85}0.38&&\cellcolor[gray]{.95}0.13&&&\cellcolor[gray]{.9}0.25&&&\cellcolor[gray]{.96}0.09&&&&&\\
        231&0.01&&&&&&&&&\cellcolor[gray]{.83}0.42&&\cellcolor[gray]{.88}\textbf{0.29}&&\cellcolor[gray]{.97}0.08&0.01&\cellcolor[gray]{.99}0.03&\cellcolor[gray]{.94}0.14&&&\cellcolor[gray]{.99}0.02&&&&&\\
        242&\cellcolor[gray]{.98}0.04&&&&&&&&&\cellcolor[gray]{.77}0.57&&0.1&\textbf{0.01}&\cellcolor[gray]{.94}0.15&0.01&\cellcolor[gray]{.99}0.02&\cellcolor[gray]{.96}0.09&&&&&&&&\\
        243&\cellcolor[gray]{.98}0.04&&&&&&&&&\cellcolor[gray]{.89}0.28&&\cellcolor[gray]{.97}0.08&&\cellcolor[gray]{.9}\textbf{0.26}&\cellcolor[gray]{.99}0.03&\cellcolor[gray]{.98}0.04&\cellcolor[gray]{.90}0.24&&&\cellcolor[gray]{.99}0.02&&&&&\\
        311&0.01&&&&&&&&&\cellcolor[gray]{.96}0.1&&\cellcolor[gray]{.99}0.02&&\cellcolor[gray]{.98}0.04&\cellcolor[gray]{.96}\textbf{0.1}&\cellcolor[gray]{.96}0.09&\cellcolor[gray]{.76}0.59&&&\cellcolor[gray]{.98}0.04&&&&&\\
        \addlinespace
        312&&&&&&&&&&\cellcolor[gray]{.99}0.03&&0.01&&\cellcolor[gray]{.99}0.02&0.01&\cellcolor[gray]{.80}\textbf{0.49}&\cellcolor[gray]{.84}0.39&&&\cellcolor[gray]{.98}0.05&&&0.01&&\\
        313&&&&&&&&&&\cellcolor[gray]{.97}0.07&&\cellcolor[gray]{.99}0.02&&\cellcolor[gray]{.99}0.03&\cellcolor[gray]{.99}0.02&\cellcolor[gray]{.92}0.19&\cellcolor[gray]{.75}\textbf{0.62}&&&\cellcolor[gray]{.98}0.04&&&&&\\
        321&\cellcolor[gray]{.99}0.03&&&&&&&&&\cellcolor[gray]{.97}0.08&&\cellcolor[gray]{.92}0.2&&\cellcolor[gray]{.95}0.13&\cellcolor[gray]{.98}0.05&0.05&\cellcolor[gray]{.87}0.32&\textbf{0}&&\cellcolor[gray]{.97}0.08&&\cellcolor[gray]{.98}0.05&&&0.01\\
        322&&&&&&&&&&\cellcolor[gray]{.99}0.02&&\cellcolor[gray]{.83}0.42&&\cellcolor[gray]{.97}0.07&&&\cellcolor[gray]{.81}0.48&&\textbf{0}&&&&&&\\
        324&0.01&&&&&&&&&\cellcolor[gray]{.98}0.05&&\cellcolor[gray]{.99}0.03&&\cellcolor[gray]{.98}0.05&\cellcolor[gray]{.99}0.03&\cellcolor[gray]{.93}0.18&\cellcolor[gray]{.83}0.43&&&\cellcolor[gray]{.93}\textbf{0.18}&&&\cellcolor[gray]{.98}0.04&&\\
        \addlinespace
        331&&&&&&&&&&&&&&&&\cellcolor[gray]{.93}0.18&&&&&\textbf{0}&\cellcolor[gray]{.94}0.16&&\cellcolor[gray]{.74}0.65&0.01\\
        411&\cellcolor[gray]{.99}0.02&&&&&&&&&\cellcolor[gray]{.99}0.02&&&&\cellcolor[gray]{.99}0.03&\cellcolor[gray]{.99}0.03&\cellcolor[gray]{.95}0.13&\cellcolor[gray]{.9}0.25&&&\cellcolor[gray]{.9}0.26&&\cellcolor[gray]{.96}\textbf{0.09}&\cellcolor[gray]{.94}0.16&\cellcolor[gray]{.99}0.02&\\
        412&&&&&&&&&&&&&&&0.01&\cellcolor[gray]{.95}0.13&\cellcolor[gray]{.98}0.05&&&\cellcolor[gray]{.88}0.3&&&\cellcolor[gray]{.80}\textbf{0.51}&&\\
        512&&&&&&&&&&&&&&\cellcolor[gray]{.98}0.05&&\cellcolor[gray]{.98}0.04&\cellcolor[gray]{.98}0.04&&&0.01&&\cellcolor[gray]{.99}0.03&\cellcolor[gray]{.99}0.03&\cellcolor[gray]{.74}\textbf{0.66}&\cellcolor[gray]{.94}0.14\\
        523&&&&&&&&&&\cellcolor[gray]{.99}0.02&&&&\cellcolor[gray]{.99}0.03&&0.01&\cellcolor[gray]{.98}0.04&&&0.01&&\cellcolor[gray]{.99}0.02&&\cellcolor[gray]{.91}0.23&\cellcolor[gray]{.74}\textbf{0.64}\\
        \bottomrule
    \end{tabularx}
\end{table*}

\subsection{Noise in datasets}

\begin{figure}[!htbp]
  \centering
  \subfigure[Predicted: Agro-forestry areas (Probability: 0.54), True label: Continuous urban fabric (Loss: 6.64).]
    {\includegraphics[width=0.48\linewidth]{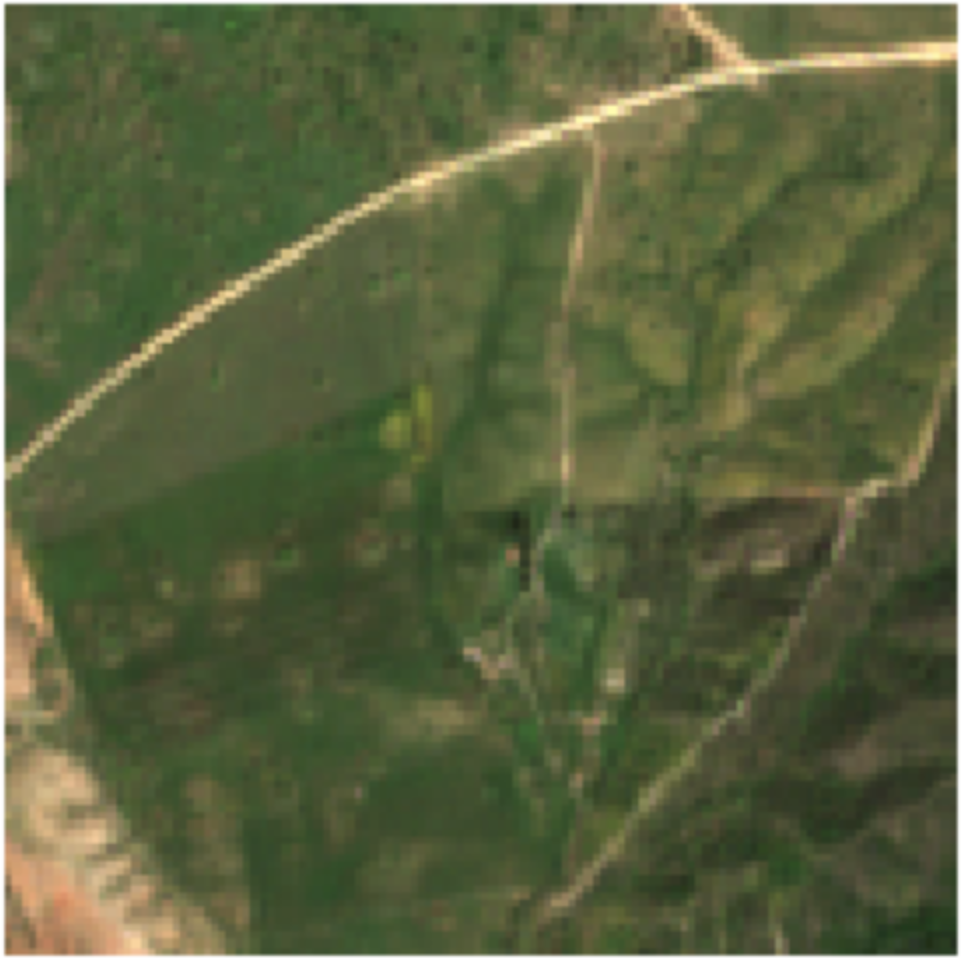}}
  \hfill
  \subfigure[Predicted: Coniferous forest (Probability: 0.78), True label: Continuous urban fabric (Loss: 6.86).]
    {\includegraphics[width=0.48\linewidth]{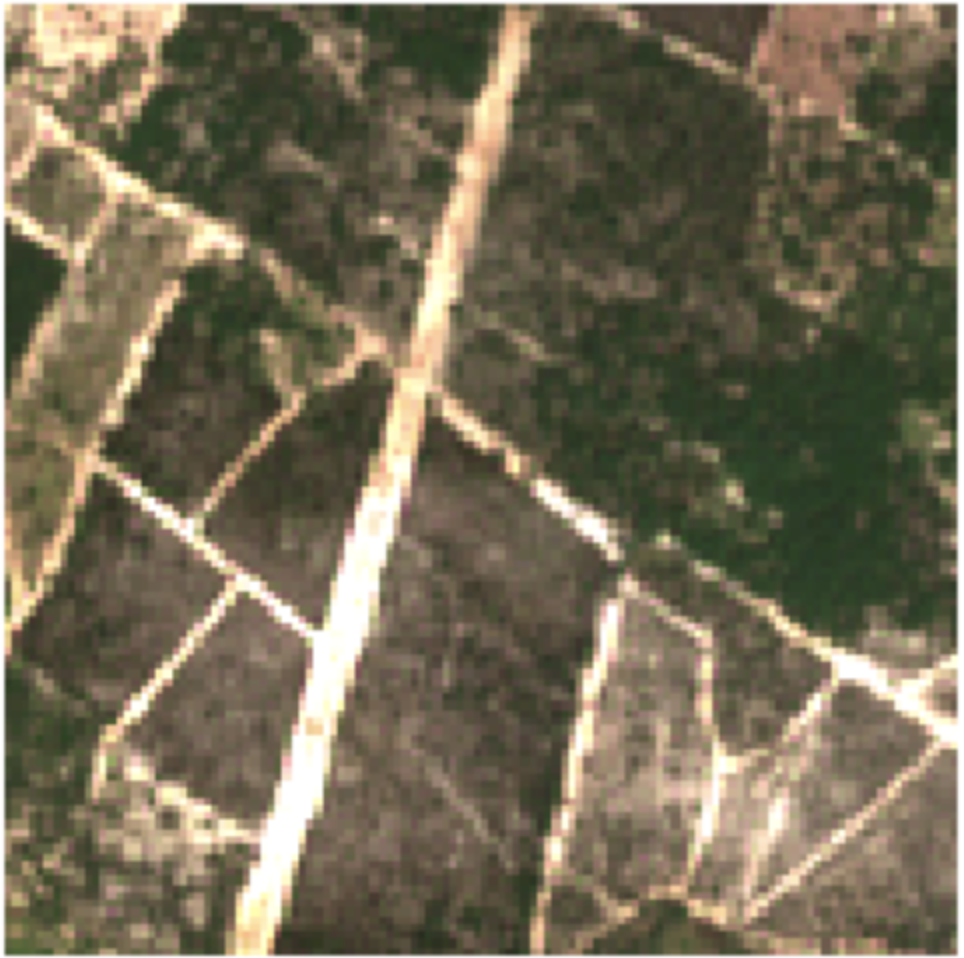}}
  \subfigure[Predicted: Agricultural areas (Probability: 0.89), True label: Artificial surfaces (Loss: 6.43).]
    {\includegraphics[width=0.48\linewidth]{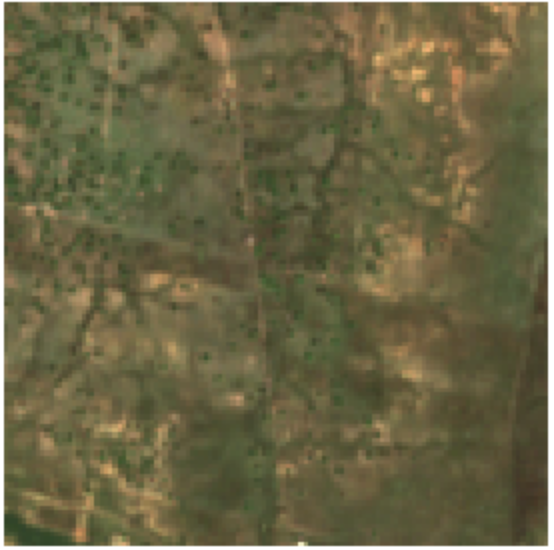}}
  \hfill
  \subfigure[Predicted: Broad-leaved forest (Probability: 0.67), True label: Continuous urban fabric (Loss: 7.65).]
    {\includegraphics[width=0.48\linewidth]{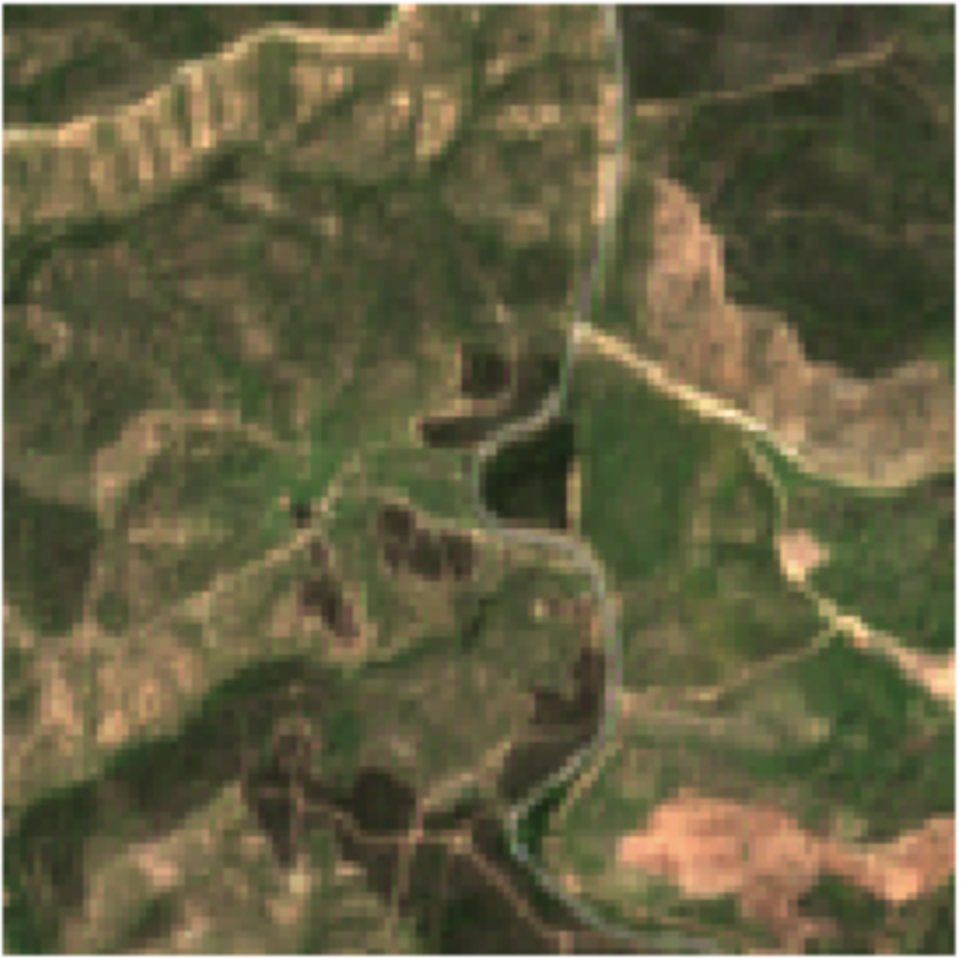}}
  \caption{Noise in classification data. Possible incorrect labels indicated by high loss values.}
  \label{fig:BigEarthNet_noise}
\end{figure}

Analysis of the classification model by viewing misclassified images with the highest loss indicates noise in the BigEarthNet data labels. Figure~\ref{fig:BigEarthNet_noise} shows one example of this, where the actual label seems to be incorrect. This method can be used further to improve the BigEarthNet dataset by correcting image labels.

\begin{figure}[!htbp]
  \centering
  \subfigure[]
    {\includegraphics[width=0.48\linewidth]{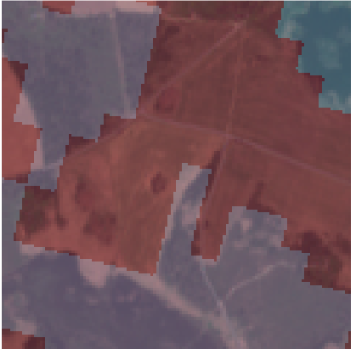}}
  \hfill
  \subfigure[]
    {\includegraphics[width=0.48\linewidth]{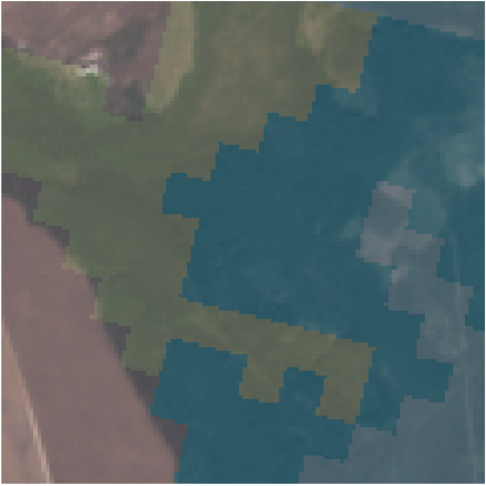}}
  \subfigure[]
    {\includegraphics[width=0.48\linewidth]{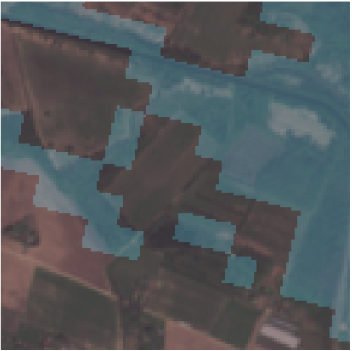}}
  \hfill
  \subfigure[]
    {\includegraphics[width=0.48\linewidth]{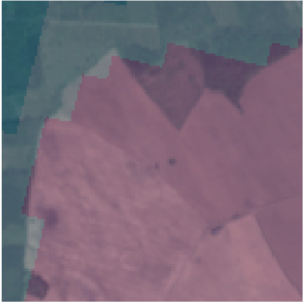}}
  \caption{CORINE Land Cover map overlaid on Sentinel-2 satellite image indicates noise in the segmentation dataset.}
  \label{fig:CORINE_noise}
\end{figure}

Combining the Sentinel-2 tile with the CORINE Land Cover map to generate a segmentation dataset also generates noise (Figure \ref{fig:CORINE_noise}). Firstly, the lower spatial resolution map is transposed to match a higher resolution satellite image. Secondly, the CORINE land cover map has an estimated accuracy of about 85\% and it omits areas less than 25 hectares and linear instances less than 100m wide.

\subsection{Discussion}

Firstly, we saw from analysing the BigEarthNet dataset how the complexity of the task increases between the three land cover classification levels. This is understandable for a hierarchical classification structure.

Training of the machine learning models was similar for all three levels of classification. In all cases the model training started with an ImageNet pretrained ResNet50 model. From there all classification models were trained 10-15 epochs, which was sufficient to see the model accuracy plateau.

Training the segmentation models started with the trained classification models and a transfer learning method was used. The classification models were used as the encoders of the U-Net like architecture with the layers frozen during the first epochs. This allowed to first focus on training the decoder side of the model. After this all layers were trained in order to retrain the encoder side as well.

The classification models showed higher results on all three levels mainly for forest, agricultural areas and water bodies. With increasing the classification level, the complexity of the task increased and many smaller classes were added, which the model was not able to correctly classify.

When we look at misclassified results by the biggest loss, we can identify images which seem to have incorrect labels in the BigEarthNet dataset. Figure \ref{fig:BigEarthNet_noise} shows examples of such images. Using the proposed classification model approach is a good way for identifying such mislabelled images for improving the dataset.

For all trained models the training was done until the accuracy metric reached a plateau and did not improve further. It can be possible to further improve the model by hyperparameter tuning and longer training, but this was sufficient for the current analysis.

Although the models show high result on some classes there are many smaller classes which show low results. One reason for this is understandably the fact that the data is imbalanced. From the results we can also see that visual distinction is an important factor as well. As we go to the 2nd and 3rd level of classification the visual distinction between the classes becomes smaller and some classes are not possible to determine on small scale images even for humans. One solution to this is to reduce the number of classes with a focus on usability in machine learning model training. This has been recently carried out for the BigEarthNet dataset in \cite{SUMBUL2020}. The third important factor is also the accuracy of training and validation datasets. The CORINE Land Cover map has been created with an aimed accuracy of 85\% which contributes to the dataset noise. Also, the lower resolution of the land cover maps means that there is even more noise as the borders between different classes are less accurate compared to satellite images. This can be seen on Figure \ref{fig:CORINE_noise}.

Considering the noise in the data described above it can be seen how the segmentation model in some cases is able to produce higher accuracy class borders. This also illustrates how the model is able to overcome the noise in the training data.

The process and machine learning models described in this work can be used for creating solutions in land monitoring and change detection. One such direction can be in monitoring changes in forest reserves, flagging deforestation logging activity. This is also part of UN Sustainable Development goals. Several further development directions have also been pointed out in the Conclusion section of this paper.

The classification ResNet model is also a good starting point for many other machine learning solutions based on satellite imagery. The model is trained on a large dataset and therefore has a good understanding of satellite image features. Depending on the problem it might only be needed to retrain the last layers of the model to solve a new task.

\section{Conclusion}
\label{sec:conclusion}
The goal of this paper was to create machine learning models for classification and segmentation of satellite imagery with the aim of improving existing land cover maps and land cover change detection.

A set of classification and segmentation models were created for satellite image classification and pixel level segmentation according to a three-level land cover category classification defined by the CORINE program. A novel BigEarthNet dataset was used, which is a public satellite imagery dataset for machine learning application. Also, a dataset was composed for training segmentation models, combining a Sentinel-2 satellite image of Estonia with the CORINE land cover map.

The results of this paper show the possibilities of using a convolutional neural network for this sort of task and underline the need for changing the land cover categories for achieving better accuracy. Also, as land cover is severely unbalanced, there is a need for class-based analysis and accuracy measurement, which was used in this work.

As an important additional result, the modified U-Net models show\-ed a capability of improving on the existing low resolution of land cover maps (Figure \ref{seg_lvl1_comparison}). The models used existing map data as input and managed to offer up improved land cover mappings compared to the data used for validation.

The use of BigEarthNet and CORINE land cover datasets for machine learning highlighted some noise in the data, which affects the results. In BigEarthNet some images seem to be mislabelled and the authors suggest a method based on classification loss can be used to find the images for label correction. With land cover maps a limitation is its accuracy of 85\%, a relatively low 100m resolution and the fact that distinctive areas under 25ha are not included on the map.

The goals of this paper were achieved and a total of six convolutional neural networks were created in order to analyse land cover classification and segmentation on three classification levels set by CORINE land cover mapping. In addition, several ideas for the improvement of the results and further research were proposed.

\subsection{Contributions of the paper}
Contributions of this paper can be summarized as follows:

\begin{itemize}
  \item A novel and very recent large-scale dataset, BigEarthNet, was used and state-of-the-art convolutional neural network architectures were applied in the research.
  \item Class based analysis of land cover classification and pixel level segmentation was carried out, highlighting the need for optimising the list of classes for machine learning purposes.
  \item An existing, manually created land cover dataset was used for machine learning model training. In visual comparison the created segmentation models showed results with higher accuracy than the dataset used for model training and validation.
  \item The research highlighted discrepancies in the BigEarthNet dataset and described a method for improving the dataset.
\end{itemize}

\subsection{Limitations and Directions for future research}

This paper presented contributions, which can be further investigated from multiple interesting perspectives. Main current limitations and directions for future research can be summarized as follows:

\begin{itemize}
  \item The accuracy of the CORINE Land Classification dataset, which was used in training and validation, was an important factor in the current results. A higher accuracy dataset will allow to increase the model accuracy and the reliability of model validation.
  \item Current work only used the red, green and blue channels of Sentinel-2 data. Including more channels allows for improving the results further. This data is available in BigEarthNet as well.
  \item Satellite data can be used for machine learning as a time series. This would allow to reinforce the land cover mapping confidence through time.
  \item The classification model approach can be used to flag and amend possible discrepancies in the BigEarthNet dataset. This can be done by viewing image classifications with high loss.
  \item The CORINE land cover classification used in this work is not optimal for machine learning based segmentation, we can see increased accuracy on some classes (forests, arable land, water bodies, for example) and we can see low results on classes which are visually less distinct. One direction for improving the results is to adjust the classes for better segmentation results. This approach has recently been carried out for the BigEarthNet dataset and a similar class selection could be used in image segmentation as well.
  \item The hierarchical structure of the CORINE land cover classification can be further utilised to improve the results. The results on the higher level can be used as a direction on the lower level.
\end{itemize}

\subsubsection*{Acknowledgement}
This research was partially supported by IT Academy and Business Information Technology programs.

\bibliographystyle{IEEEtran}
\bibliography{References}
\vskip 0pt plus -1fil
\begin{IEEEbiography}[{\includegraphics[width=1in,height=1.25in,clip,keepaspectratio]{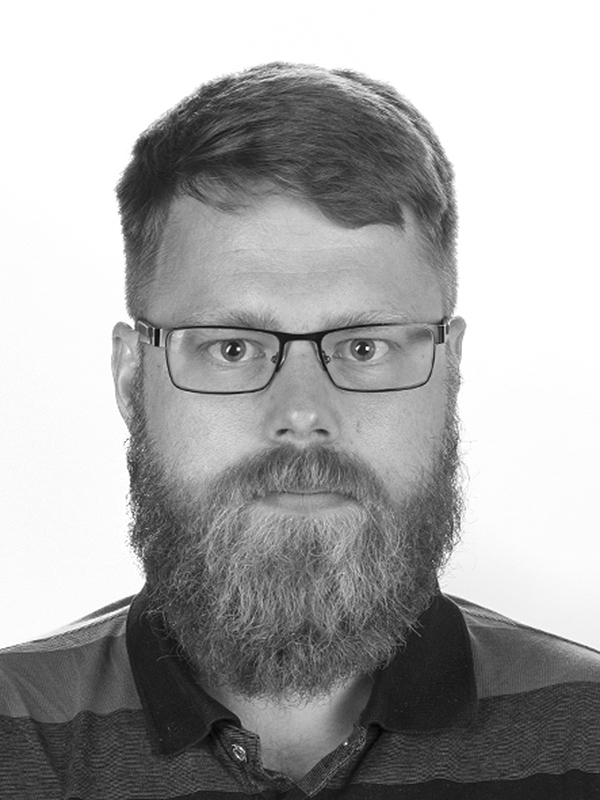}}]{Priit Ulmas} received the M.Sc. degree in civil engineering from Tallinn University of Technology, Tallinn, Estonia in 2013 and the second M.Sc. in business information technology from Tallinn University of Technology, Tallinn, Estonia in 2020.

He has over 10 years of industry engineering experience. Having previously worked on virtual design and construction and project management, he is currently working as a Machine Learning Engineer, developing automation solutions for manufacturing industry. His current research interests include image processing, machine learning, and computer vision.

\end{IEEEbiography}
\vskip 0pt plus -1fil

\begin{IEEEbiography}[{\includegraphics[width=1in,height=1.25in,clip,keepaspectratio]{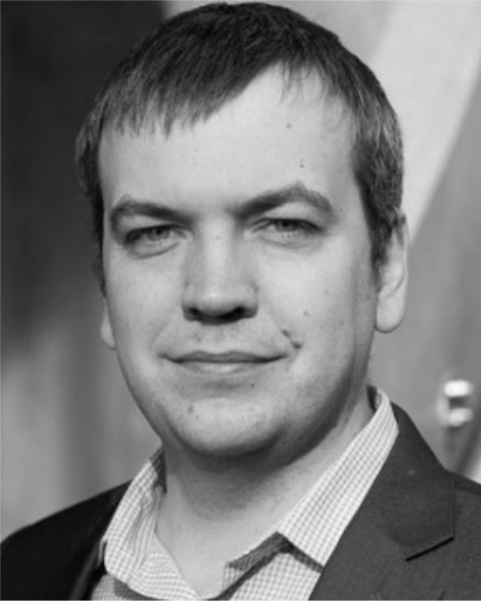}}]{Innar Liiv} received the Ph.D. degree from Tallinn University of Technology, Estonia.
He was a Cyber Studies Visiting Research Fellow with the University of Oxford, from 2016 to 2017, a Visiting Scholar with Stanford University, in 2015, and a Postdoctoral Visiting Researcher with the Georgia Institute of Technology, in 2009. He is currently an Associate Professor of data science with the Tallinn University of Technology and a Research Associate
with the Centre for Technology and Global Affairs, Oxford University. His research interests include e-government and data science, information visualization, and big data technology transfer to industrial and governmental applications. He is an IEEE Senior Member.
\end{IEEEbiography}
\EOD
\end{document}